\documentclass{ieeeaccess}
\usepackage{cite}
\usepackage{amsmath,amssymb,amsfonts}
\usepackage{algorithm, algpseudocode}
\usepackage{graphicx}
\usepackage{textcomp}
\usepackage{varwidth}

\usepackage{soul}
\DeclareRobustCommand{\hlnew}[1]{{\sethlcolor{white}\hl{#1}}}

\usepackage[nolist]{acronym}

\newacro{lfd}[LfD]{Learning from Demonstration}
\newacro{spd}[SPD]{Symmetric Positive Definite}
\newacro{rm}[RM]{Riemannian Manifold}
\newacro{uq}[UQ]{Unit Quaternion}
\newacro{dtw}[DTW]{Dynamic Time Warping}
\newacro{gmm}[GMM]{Gaussian Mixture Model}
\newacro{tpgmm}[TP-GMM]{Task-Parameterized Gaussian Mixture Model}
\newacro{gmr}[GMR]{Gaussian Mixture Regression}
\newacro{dmp}[DMP]{dynamic movement primitives}
\newacro{led}[LEd]{Log Euclidean distance}
\newacro{lqd}[LQd]{Log Quaternion distance}
\newacro{kmp}[KMP]{Kernelized Movement Primitive}
\newacro{dof}[DoF]{Degrees-of-Freedom}
\newacro{fdm}[FDM]{Fast Diffeomorphic Matching}
\newacro{ros}[ROS]{Robot Operating System}
\newacro{emg}[EMG]{Electromyography}
\newacro{pih}[PiH]{Peg-in-Hole}

\newcommand{\bm}[1]{\boldsymbol{\mathbf{#1}}}

\renewcommand{\thetable}{\Roman{table}}

\newcommand{\mathcalbold}[1]{\boldsymbol{\mathcal{#1}}}
\newcommand{\mathcalbb}[1]{\boldsymbol{\mathbb{#1}}}
\newcommand{\mathcalfk}[1]{\boldsymbol{\mathfrak{#1}}}

\newcommand{\Logg}{\text{Log}_{\bm{g}}}

\newcommand{\Log}{\text{Log}}

\newcommand{\Expg}{\text{Exp}_{\bm{g}}}

\def\mfkpo{$\mathcalfk{p}_1$}
\def\mfkpt{$\mathcalfk{p}_2$}
\def\mfkph{$\mathcalfk{p}_3$}

\newcommand{\trsp}{{^{\top}}}

\newcommand{\figref}[1]{Fig.~\hyperref[#1]{\ref*{#1}}}
\newcommand{\figsref}[1]{Figures~\hyperref[#1]{\ref*{#1}}}
\newcommand{\Figref}[1]{Figure~\hyperref[#1]{\ref*{#1}}}

\newcommand{\tabref}[1]{Tab.~\hyperref[#1]{\ref*{#1}}}
\newcommand{\secref}[1]{Section~\hyperref[#1]{\ref*{#1}}}
\newcommand{\algoref}[1]{Algorithm~\hyperref[#1]{\ref*{#1}}}

\newcommand{\wrt} {\textit{w.r.t.}} %
\newcommand{\eg} {{e.g.,}~} %
\newcommand{\ie} {{i.e.,}~} %
\newcommand{\etal}{\MakeLowercase{\textit{et al.\ }}}

\newlength{\Oldarrayrulewidth}

\definecolor{darkgreen}{rgb}{0.0,0.49,0.19}

\newcommand{\panda}{Franka Emika Panda}

\relax
\DeclareFontFamily{U}{eur}{\skewchar\font'177}
\DeclareFontShape{U}{eur}{m}{n}{%
	<-6> eurm5 <6-8> eurm7 <8-> eurm10}{}
\DeclareFontShape{U}{eur}{b}{n}{%
	<-6> eurb5 <6-8> eurb7 <8-> eurb10}{}
\DeclareSymbolFont{ugrf@m}{U}{eur}{m}{n}
\SetSymbolFont{ugrf@m}{bold}{U}{eur}{b}{n}
\DeclareMathSymbol{\upomega}{\mathord}{ugrf@m}{"21}
\graphicspath{{figs/}}

\def\BibTeX{{\rm B\kern-.05em{\sc i\kern-.025em b}\kern-.08em
		T\kern-.1667em\lower.7ex\hbox{E}\kern-.125emX}}
\begin{document}
\history{Date of publication xxxx 00, 0000, date of current version xxxx 00, 0000.}
\doi{10.1109/ACCESS.2017.DOI}

\title{Learning Deep Robotic Skills on Riemannian manifolds}
\author{\uppercase{Weitao Wang}\authorrefmark{1},
	\uppercase{Matteo Saveriano}\authorrefmark{2}, \IEEEmembership{Member, IEEE}, and \uppercase{Fares J. Abu-Dakka}\authorrefmark{1}, \IEEEmembership{Member, IEEE}}
\address[1]{Intelligent Robotics Group, Department of Electrical Engineering and Automation (EEA), Aalto University,	Espoo, Finland (e-mail: \{weitao.wang,fares.abu-dakka\}@aalto.fi)}
\address[2]{Department of Industrial Engineering (DII), University of Trento, Trento, 38123, Italy (e-mail: matteo.saveriano@unitn.it)}
\tfootnote{Part of the research presented in this work has been conducted when M. Saveriano was at the Department of Computer Science, University of Innsbruck, Innsbruck, Austria. This work has been partially supported by CHIST-ERA project IPALM (Academy of Finland decision 326304).}

\markboth
{Wang \headeretal: RiemannianFlow}
{Wang \headeretal: RiemannianFlow}

\corresp{Corresponding author: Fares~J.~Abu-Dakka (e-mail: fares.abu-dakka@aalto.fi).}

\begin{abstract}
In this paper, we propose RiemannianFlow, a deep generative model that allows robots to learn complex and stable skills evolving on Riemannian manifolds. Examples of Riemannian data in robotics include stiffness (symmetric and positive definite matrix \hlnew{(SPD)}) and orientation (unit quaternion \hlnew{(UQ)}) trajectories.  For Riemannian data, unlike Euclidean ones, different dimensions are interconnected by geometric constraints which have to be properly considered during the learning process. Using distance preserving mappings, our approach transfers the data between their original manifold and the tangent space, realizing the removing and re-fulfilling of the geometric constraints. This allows to extend existing frameworks to learn stable skills from Riemannian data while guaranteeing the stability of the learning results. The ability of RiemannianFlow to learn various data patterns and the stability of the learned models are experimentally shown on a dataset of manifold motions. Further, we analyze from different perspectives the robustness of the model with different hyperparameter combinations. 
It turns out that the model's stability is not affected by different hyperparameters, a proper combination of the  hyperparameters leads to a significant improvement (up to 27.6\%) of the model accuracy. Last, we show the effectiveness of RiemannianFlow in a real peg-in-hole \hlnew{(PiH)} task where we need to generate stable and consistent position and orientation trajectories for the robot starting from different initial poses.
\end{abstract}


\titlepgskip=-15pt

\maketitle

\section{Introduction}
\label{sec:introduction}
\begin{figure}[t]
	\centering
	\includegraphics[width=\linewidth]{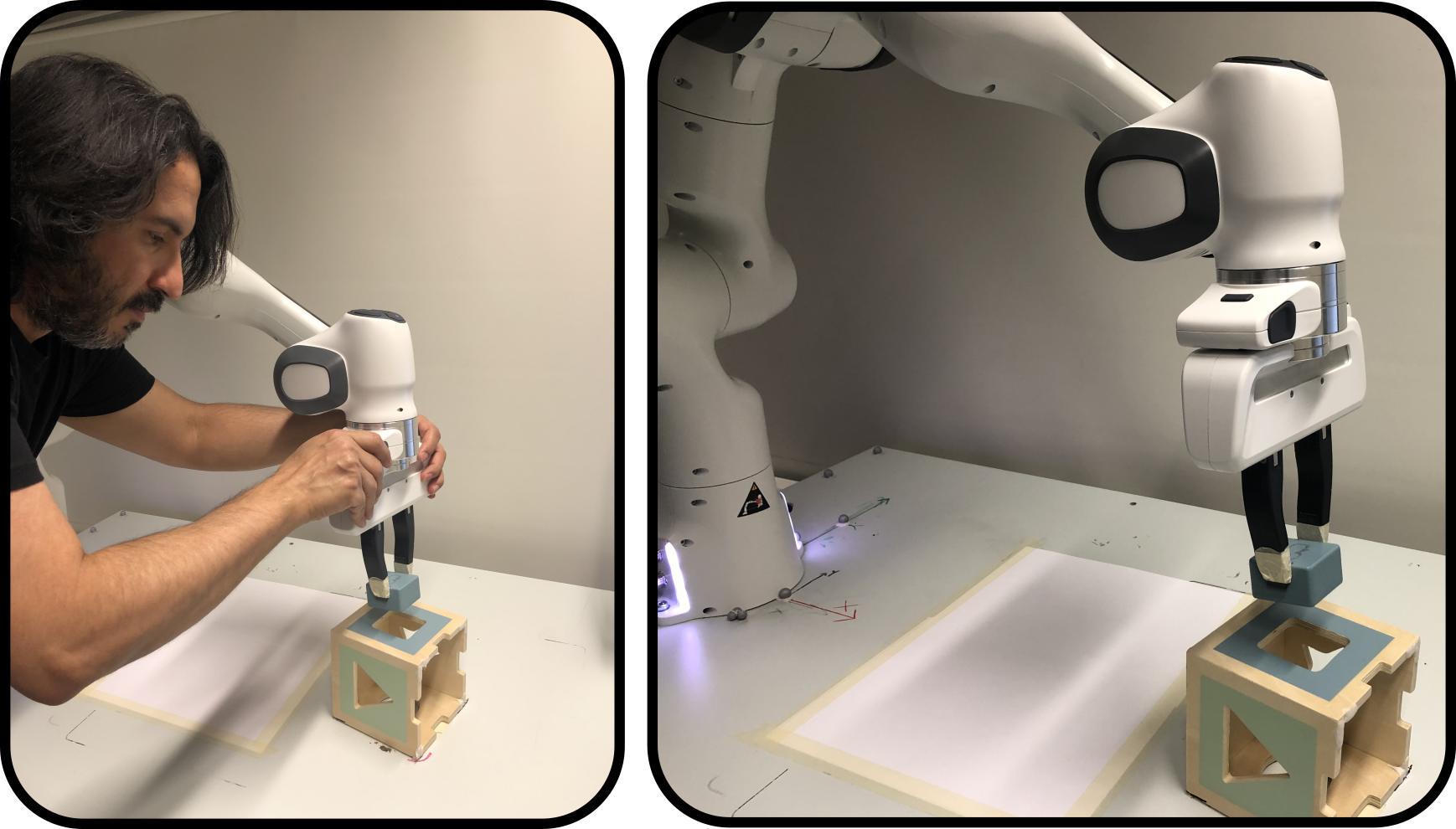} 
	\caption{\emph{Left}: a human operator teaches a robot how to perform a \ac{pih} task. \emph{Right}: a snapshot showing the \ac{pih} task executed by a \panda~robot.}
	\label{fig:LfD}
\end{figure}

\IEEEPARstart{R}{obots} operating in everyday environments nowadays are required not only to follow some rigid movements, but also to accomplish complex physical interactions with the environment, including tools, humans, and/or other robots. During these activities, the reference position, stiffness, and orientation are changing along with the procedure, which makes it extremely time-consuming to implement with traditional hard-programming methods. In such a situation, \ac{lfd} \cite{ravichandar2020recent} is considered as a more efficient way for robots to acquire new abilities, as even people with no robotics knowledge are able to give demonstrations teaching robots their own specialties (see Fig.\ref{fig:LfD}). The learning-based solution proposed in this work can be directly exploited by robotics experts and non-experts to accomplish desired tasks without complex programming, which will accelerate the transition from ideas to products in manufacturing and save much cost at the same time.

The Trajectory-tracking task is an active research topic in robotics, which is indeed a fundamental component in a \ac{lfd} system. In these path-following tasks, the main focus is to generate a robust and precise route \wrt\footnote{\hlnew{\wrt stands for \textit{with respect to}}} the position and orientation\hlnew{\mbox{\cite{VidakovicJosip2020Artl,XuSheng2019Rttc}}}. However, to achieve more complicated motions, different kinds of control information like force and stiffness~\cite{abu2018force} are regarded as objects to be learned from demonstrations. Unlike simple data, \eg data of position or joint angles where \acp{dof} are independent, some data are represented by special formats. For instance, when we handle stiffness, inertia, or sensory data organized as covariance features, \ac{spd} matrices are encountered, while \acp{uq} are used to represent orientations. For such data, \acp{dof} are related by particular constraints arising from the structure of the underlying space, which makes the learning problem more difficult as the learner has also to fulfill those constraints.

In this paper, we propose a geometry-aware approach to correctly handle complex data structures and their underlying constraints. In particular, we focus on data evolving on \acp{rm}. As handling geometric constraints directly on the \ac{rm} is complex, we exploit a distance preserving mapping to move the data from a manifold to a local tangent space. As the tangent space is a finite dimensional linear space, it is isomorphic to the Euclidean space and tools from linear algebra can be applied freely on the transformed data. On the tangent space, the transformed data are considered as generated from a stochastic nonlinear dynamical system whose dynamics is learned using the ImitiationFlow approach~\cite{UrainJulen2020ILDS}. The learned model is used at run-time to retrieve the data on the tangent space which are then projected back to the original \ac{rm} using a reverse mapping. The inverse mapping automatically imposes constraints of the geometric structure of the manifold on the tangent space motion. {A similar approach is used in~\cite{urain2021learning} to learn stable \ac{uq} motions. The main difference is that~\cite{urain2021learning} works on Lie groups like \ac{uq}, while our approach is more general as it works also on manifolds like \ac{spd} matrices that are not a Lie group.} In our experiments, the proposed RiemannianFlow is proved to be able to accurately learn stable dynamical systems evolving on different, possibly high dimensional \acp{rm}. To summarize, our contribution is three-fold:
\begin{itemize}
    \item We propose a geometry-aware approach to learn stable robotic skills on \acp{rm}.
    \item We show that our approach allows to straightforwardly extend existing approaches, including modern deep learning techniques, to learn from manifold data.
    \item We compare our approach with two baselines and two state-of-the-art approaches on a public benchmark.
\end{itemize}

The rest of the paper is organized as follows. Section~\ref{sec:related} presents the related work and highlights similarities and differences with our method. In Sec.~\ref{sec:background}, basic knowledge of \ac{rm} and the \ac{lfd} model used in this work are introduced. Section~\ref{sec:method} presents the proposed approach. We compare RiemannianFlow with two baseline and two state-of-the-art approaches in Sec.~\ref{sec:experiment} and show an experiment in a typical industrial task. Section~\ref{sec:conclusion} states the conclusions and proposes further research directions.

\section{Related work}\label{sec:related}
For robot-environment interaction, robots need to have a sophisticated adaptation of their end-effector pose and stiffness. The importance of controlling pose and stiffness for the successful accomplishment of robotic tasks makes them fundamental research topics in the field of robot manipulation.

For the stiffness, an intuitive method was provided by~\cite{chen2020neural} to generate the stiffness directly from the limb postures and surface \ac{emg} signals of the human demonstrator. \hlnew{Authors in \mbox{\cite{peternel2018robot}} used human demonstration along with \mbox{\ac{emg}} to learn stiffness from human muscle activity measurements.} The learning framework proposed in~\cite{abu2018force} used the measured trajectories and interaction forces to derive a full stiffness matrices profile encoded in a probabilistic model to be executed later with unseen situations. Kronander and Billard~\cite{KronanderKlas2014LCMt} exploited the variability in the demonstrations, injected by shacking the robot during the motion, to learn a variable stiffness profile where high variance corresponds to low stiffness (less accurate tracking). \acp{kmp}~\cite{huang2019} were exploited in~\cite{abudakka2021probabilistic} to generalize variable stiffness profiles. Jaquier~\etal~\cite{jaquier2021geometry} proposed a \ac{gmm}-based framework to learn \ac{spd} profiles with a particular focus on robot manipulability.

For the orientation, early work~\cite{ silverio2015learning} did not consider geometric constraints---unit norm for \acp{uq} or orthogonality for rotation matrices---when they were learning the orientation data. Instead, they modified the generated trajectory at run-time to fulfill the constraints, causing deviations from the demonstrated motion. To remedy this issue, the work in~\cite{Ude2014, AbuDakka2015} extended the classical \ac{dmp} formulation~\cite{saveriano2021dynamic} to properly handle orientation data. The stability of the obtained approach is shown in~\cite{saveriano2019merging} for \acp{uq}.
Abu-Dakka~\etal~\cite{abudakka2021periodic} extended periodic \acp{dmp} to encode periodic orientation patterns. \acp{gmm} and \acp{tpgmm} were extended to describe the distribution of \acp{uq} in \cite{kim2017gaussian} and \etal~\cite{Zeestraten2017} respectively. \hlnew{Rozo and Dave~\mbox{\cite{Rozo2021OrientationPM}} proposed a Riemannian extension of the probabilistic movement primitives framework.} \acp{kmp} were also extended to represent orientation trajectories by exploiting the mappings between \ac{uq} manifold and its tangent space~\cite{huang2020toward}.

RiemannianFlow exploits the idea of projecting data in the tangent space to remove geometric constraints and then re-fulfilling them by projecting the generated data back to the manifold. Working in the tangent space allows to use standard tools from Euclidean geometry to encode the demonstrations and fit a stable dynamics in the tangent space to ensure convergence to a given target on the manifold. Moreover, it makes the formulation relatively general and easy to adapt to different manifolds. Indeed, the major change is to use the proper projection operations which are manifold dependent. It is worth mentioning that working in the tangent space is a local approach that is exact only if the data belong to the same chart on the manifold. This problem affects \acp{uq}, but not \ac{spd} matrices \cite{PENNEC202075}.  As shown in~\cite{Calinon20RAM}, if quaternion data span multiple charts then using a single tangent space introduces significant approximation errors. Instead of using multiple tangent spaces as in~\cite{Calinon20RAM}, which requires clustering of the data, we use a simple pre-processing step to ensure that quaternion data belong to the same hemisphere (see Sec.~\ref{subsec:preprocessing}).

\section{Background}
\label{sec:background}
This section briefly describes the geometric structure of \ac{spd}, \ac{uq} manifolds and the key aspects of ImitationFlow~\cite{UrainJulen2020ILDS}.

\subsection{Distance preserving transformations on \hlnew{rm}\lowercase{s}}
In this paper, we focus mainly on two \acp{rm}, namely the $d \times d$ \ac{spd} matrices $\mathcalbold{S}^{+}_d$ and the unit-sphere $\mathcalbold{S}^3$ (\acp{uq}). As already mentioned, the RiemannianFlow works by moving manifold data into the local tangent space $\mathcalbold{TM}$ and back to the manifold $\mathcalbold{M}$. We are also interested in preserving the convergence of the learned motion to a given point on the manifold, namely the goal $\bm{g} \in \mathcalbold{M}$. Therefore, we consider the tangent space at the goal $\mathcalbold{T}_{\bm{g}} \mathcalbold{M}$ and project our data on it.  For this, we need to define the logarithmic mapping function $\Logg(\bm{p}): \mathcalbold{M} \rightarrow \mathcalbold{T}_{\bm{g}} \mathcalbold{M}$ which moves a point $\bm{p}$ from the manifold to $\mathcalfk{p}$ on the tangent space of $\bm{g}$, along the projection of the geodesic (the shortest curve on the \ac{rm}) between $\bm{p}$ and $\bm{g}$. In this way, data on the \ac{rm} can be transferred to the tangent space to be manipulated freely. After the learning procedure, the exponential mapping function $\Expg (\mathcalfk{p}): \mathcalbold{T}_{\bm{g}}\mathcalbold{M} \rightarrow \mathcalbold{M}$, which is the inverse of the logarithmic mapping, can transfer the data back to the \ac{rm}.

For the manifold $\mathcalbold{S}^{+}_d$, $\bm{p}$ and $\bm{g}$ are represented by \ac{spd} matrices and $\mathcalfk{p}$ is described by a symmetric matrix, these 2 mappings can be computed as~\cite{PENNEC202075}:
\begin{eqnarray}
	&\Logg(\bm{p}) = \bm{g}^{\frac{1}{2}} \mathrm{logm}(\bm{g}^{-\frac{1}{2}}\bm{p}\bm{g}^{-\frac{1}{2}}) \bm{g}^{\frac{1}{2}},
	\label{spd_log} \\
	&\Expg (\mathcalfk{p}) = \bm{g}^{\frac{1}{2}} \mathrm{expm}(\bm{g}^{-\frac{1}{2}} \mathcalfk{p} \bm{g}^{-\frac{1}{2}}) \bm{g}^{\frac{1}{2}},
	\label{spd_exp}
\end{eqnarray}
where $\mathrm{logm}(\cdot)$ and $\mathrm{expm}(\cdot)$ are the matrix logarithm and exponential respectively.
Last, we use Mandel's notation to transform the symmetric matrix $\mathcalfk{p}$ to a vector $\mathrm{vec}(\mathcalfk{p})$ and vice versa. Therefore, the whole process realizes the conversion between \ac{spd} matrices and vectors.

For the manifold $\mathcalbold{S}^3$, $\bm{p}$ is represented by a \ac{uq}, where $\bm{p}= \nu + \bm{u}$, and $\mathcalfk{p}$ is described by a 3-dimensional vector. Therefore, the logarithmic and exponential mappings \cite{Ude2014} are: 
\begin{equation}
	\Logg(\bm{p}) = \Log(\bm{p}*\bm{\bar{g}}) =  
	\begin{cases}
		\arccos(\nu) \frac{\bm{u}}{||\bm{u}||}, & \bm{u}\neq \bm{0} \\
		[0\;\; 0\;\; 0]{\trsp}, & \text{otherwise}.
	\end{cases} \label{uq_log}
\end{equation}
\begin{equation}
	\Expg(\mathcalfk{p}) =
	\begin{cases}
		\left[\cos(||\mathcalfk{p}||)+\sin(||\mathcalfk{p}||)\frac{\mathcalfk{p}}{||\mathcalfk{p}||}\right] *\bm{g}, & \mathcalfk{p}\neq \bm{0} \\
		\left[1+[0\;\; 0 \,\;0]{\trsp}\right] *\bm{g}, & \text{otherwise}.
	\end{cases}
	\label{uq_exp}
\end{equation}
where $*$ denotes \acp{uq} multiplication as in \cite{Ude2014}.

\subsection{Flow-based dynamic model}
\label{subsec:imitation_flow}
ImitationFlow~\cite{UrainJulen2020ILDS} assumes that observations are sampled from an unknown, arbitrary complex distribution $\mathfrak{p} \thicksim \psi(\mathcalfk{p})$ that is related to a latent, known distribution $\mathfrak{q} \thicksim \pi(\mathcalfk{q})$ through a diffeomorphic---continuous, bijective, and with continuous derivative---map ${\mathrm{b}(\cdot): \mathcalbb{R}^d \rightarrow \mathcalbb{R}^d}$. Specifically, having the latent distribution $\mathfrak{q} \thicksim \pi(\mathcalfk{q})$, a different distribution $\mathfrak{p} \thicksim \psi(\mathcalfk{p})$ is acquired by $\mathcalfk{p} = \mathrm{b}(\mathcalfk{q})$.

In a \ac{lfd} setting, observations of the complex distribution $\psi(\mathcalfk{p})$ are given in the form of expert demonstrations and can be used to learn the diffeomorphism $\mathrm{b}(\cdot)$, \eg using a Normalizing Flow~\cite{rezende2015variational}.
To enforce the stability of the learned motion, ImitationFlow assumes that the transition model in the latent space follows a known stable dynamics. The simplest stable stochastic dynamics is given by
\begin{equation}
	\dot{\mathcalfk{q}}(t) = V_\phi \mathcalfk{q}(t) + F_\phi \bm{\beta}(t),
	\label{eq:stoc_dynamics}
\end{equation}
where $\mathcalfk{q} \in \mathcalbb{R}^d$ represents the state and $\bm{\beta}:\mathcalbb{R} \rightarrow \mathcalbb{R}^d $ (called Brownian motion or Wiener process) represents noise and variability in the demonstrations. Matrices $V_\phi$ and $F_\phi$ are parameterized by the set of parameters $\phi$. To ensure stability of the base dynamics, $\phi$ are constrained
such that the eigenvalues of $V_{\phi}$ have negative real part. The output of \eqref{eq:stoc_dynamics} is
\begin{equation}
	\mathcalfk{p} = \mathrm{b}_{\theta}(\mathcalfk{q}),
	\label{eq:diff_map}\end{equation}
where $\mathrm{b}_{\theta}$ is the diffeomorphic mapping parameterized by $\theta$. The free parameters of the model are $\theta$ and $\phi$ which can be learned from the given demonstrations. As shown in~\cite{UrainJulen2020ILDS}, the observation space dynamics obtained by differentiating~\eqref{eq:diff_map} is stable if the latent space dynamics is stable. 

The presented approach assumes that both the latent and the observation space are isomorphic to an Euclidean space. In the following section, we are discussing how to extend flow-based models to \acp{rm}. 

\section{Proposed approach}
\label{sec:method}
The problem of \acp{lfd} on stiffness or orientation profiles is to properly consider the geometric constraints imposed on these complex data structures. To tackle it, a geometry-aware approach is proposed considering that geometric constraints ``disappear'' when data are transferred from the \ac{rm} to the local tangent space. Therefore, once the data are transferred to the tangent space, standard tools can be used to learn the demonstrated patters. We choose to learn a stable trajectory in the tangent space by applying a diffeomorphic transformation on a linear and stable dynamics. To learn the diffeomorphism, we use the deep generative model proposed in~\cite{UrainJulen2020ILDS}. Last, we apply the the reverse mapping to project the generated data back to the \ac{rm} which imposes geometric constraints on the generated data. An in-depth description of our RiemannianFlow approach is given in the rest of this section.

\begin{figure}[!t]
	\centering
	\def\svgwidth{\linewidth}
	{\fontsize{8}{8}
		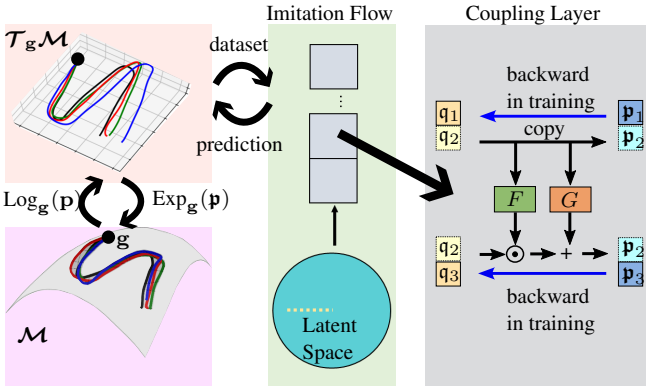}
	\caption{A pipeline indicates the whole process of the methods. Note: For each coupling layer, the second dimension is either in the upper group or in the lower group, and between each coupling layer there is a swap among dimensions to enlarge the capacity. For the 2 transformation functions: ${F, G: \mathcalbb{R}^d \rightarrow \mathcalbb{R}^d}$, they are implemented by multi-layer perceptron.}
	\label{fig:diagram}
\end{figure}

\subsection{Acquire training data from demonstration}\label{subsec:preprocessing}
We want to learn the diffeomorphic mapping $\mathrm{b}_{\theta}$, parameterized by $\theta$, that maps the trajectory of a simple dynamics into a more complex, desired one. As commonly done in \acp{lfd}, we assume that a set of $N \geq 1$ demonstrations of the same skill is given, \ie $\mathcal{D} = \{ \mathcal{D}_1,  \mathcal{D}_2, \ldots, \mathcal{D}_N \}$, where each demonstration $\mathcal{D}_n$ contains $M$ points on a \ac{rm}, \ie $\mathcal{D}_n = \{\bm{p}_1, \bm{p}_2, \ldots,\bm{p}_M\}$, and each point $\bm{p}_m \in \mathcalbold{S}^+_d$ or $\bm{p}_m \in \mathcalbold{S}^3$, $m=1,\ldots,M$. For simplicity, the sampling time $\delta t$ is assumed to be the same for all the demonstrations. Since we are interested in learning converging motions, we also assume that all the demonstrations converge to the same goal, \ie  $\bm{p}_M = \bm{g},\,\forall \mathcal{D}_n$.

The first pre-processing step, needed only if $\bm{p}_m \in \mathcalbold{S}^3$, is to check whether the dot product $\bm{p}_m \cdot \bm{p}_{m+1} > 0$. In case the dot product is negative, we substitute $\bm{p}_{m+1}$ with $-\bm{p}_{m+1}$ to prevent discontinuities in the quaternion trajectory. As a result of this step, we ensure that the entire quaternion trajectory is contained in a single chart (hemisphere) where logarithmic and exponential mappings are bijective. It is worth mentioning that a similar step is not needed for $\mathcalbold{S}^+_d$ since logarithmic and exponential maps are bijective everywhere inside the \ac{spd} cone. After this check, we project all the points in the tangent space by means of the logarithmic mapping $\Logg(\bm{p}_m)$, which is defined in~\eqref{spd_log} for \ac{spd} and in~\eqref{uq_log} for \ac{uq}. Only for \ac{spd} manifold, we vectorize the data on the tangent space using Mandel's notation. The result is the new set of demonstrations
with the same structure of the original ones but on the tangent space.
It is worth noticing that, having placed the tangent space in the last point of the demonstrations, it holds that $\mathcalfk{p}_M = \Logg(\bm{p}_M) = \Logg(\bm{g}) = \bm{0}$. 

\subsection{Deep stable dynamics on \hlnew{RM}\lowercase{s}}\label{subsec:deep_rm}
Using the approach presented in the previous section we transform the given demonstrations into a new set of demonstrations $\mathcal{T}$ containing points in the tangent space. Given the tangent space is a finite dimensional linear space, where points can be vectorized without loss of information, it is possible to exploit existing approaches to learn a stable dynamics in the tangent space. Since our targets manifolds contain complex objects like stiffness or orientation of robots, not only the  precision of the learning results matters, but also the robustness. Therefore, a generative model based on probability is chosen for its advantage of generality. In particular, we adopt a flow-based model (Sec.~\ref{subsec:imitation_flow}) to fit a stable stochastic dynamic in the tangent space. The learned model can be directly used to generate stable tangent space trajectories, as shown in Fig.~\ref{fig:4trajectories}.

To generate a Riemannian trajectory, we proceed as follows. Given an initial point  $\bm{p}_1 \in \mathcalbold{S}^+_d$ (or $\bm{p}_1 \in \mathcalbold{S}^3$) and a desired goal point $\bm{g} \in \mathcalbold{S}^+_d$ (or $\bm{g} \in \mathcalbold{S}^3$), we use the logarithmic mapping to project $\bm{p}_1$ onto the tangent space centered at $\bm{g}$, obtaining $\mathcalfk{p}_1 = \Logg(\bm{p}_1)$. Then, we pass $\mathcalfk{p}_1$ through the learned tangent space dynamic and numerically integrate the generated velocity to obtain  $\mathcalfk{p}_2$. Last, we project back the point onto the manifold by means of the exponential mapping, obtaining $\bm{p}_2 = \Expg(\mathcalfk{p}_2)$. The last step is fundamental as it re-imposes constraints on the generated point preserving the geometric structure of the manifold. We repeat until the goal is reached. To check for convergence, we recall that, having placed the tangent space in the goal, the tangent space trajectory should converge to zero. Therefore, we simply stop the procedure when the generated point $\Vert \mathcalfk{p}_m \Vert < \xi \approx 0$, where $\Vert \cdot \Vert$ is the Euclidean norm and $\xi$ is a user defined threshold. The proposed approach to generate stable motions on \acp{rm} is summarized in Algorithm~\ref{alg:riem_flow} and can be visualized in Fig.~\ref{fig:diagram}.
\begin{algorithm}[t]
	\caption{\sc{RiemannianFlow}}
	\label{alg:riem_flow}
	\begin{algorithmic}[1]
		\State Compute training data
		\begin{itemize}
			\item[--] Collect demonstrations $\mathcal{D} = \{ \mathcal{D}_1,  \mathcal{D}_2, \ldots, \mathcal{D}_N \}$ of Riemannian data $\mathcal{D}_n = \{\bm{p}_1, \bm{p}_2, \ldots,\bm{p}_M\}$
			\item[--] Define the goal $\bm{g}$, \eg the last point of each $\mathcal{D}_n$. 
			\item[--] Check if $\mathcalbold{S}^3$ data are in the same chart: $\forall \bm{p}_m \in \mathcal{D}_n \rightarrow \bm{p}_m \cdot \bm{p}_{m+1} > 0$. If not, substitute $\bm{p}_{m+1}$ with $-\bm{p}_{m+1}$
			\item[--] Project all the points to the tangent space using
			\item[] the logarithmic map:
			\item[] $\forall \bm{p}_m \in \mathcal{D}_n \rightarrow \mathcalfk{p}_m = \Logg(\bm{p}_m)$.
			\item[] Vectorize $\mathcalbold{S}^+_d$ data (\eg using Mandel's notation)
			\item[--] Save tangent space data into the set of demonstrations $\mathcal{T}= \{ \mathcal{T}_1,  \mathcal{T}_2, \ldots, \mathcal{T}_N \}$
		\end{itemize}
		\State Learn a stable motion in tangent space, \eg with ImitationFlow~\cite{UrainJulen2020ILDS}, using $\mathcal{T}$ as training data. 
		\State Generate Riemannian trajectories
	\end{algorithmic}
	\hspace{2.7em}%
	\begin{varwidth}{\linewidth}
		\begin{algorithmic}
			\State $\mathcalfk{p}_m \gets \Logg(\bm{p}_1)$
			\While{$\Vert \mathcalfk{p}_m \Vert \geq \xi$}
			\State {$\dot{\mathcalfk{p}}_{m}  \gets $ \sc{ImitationFlow}($\mathcalfk{p}_m$)}
			\State $\bm{p}_{m} \gets \Expg(\dot{\mathcalfk{p}}_{m}\delta t)$
			\EndWhile
		\end{algorithmic}	
	\end{varwidth}
\end{algorithm}

\section{Experimental evaluation}
\label{sec:experiment}

In this section, we test RiemannianFlow both in simulations and a real experiment. For the simulations, we augment a popular dataset~\cite{khansari2011learning} with \ac{uq} and \ac{spd} data. For the experiment, we use a $7$ degrees-of-freedom Franka Emika Panda manipulator to perform a \ac{pih} task.

\subsection{Dataset creation}\label{subsec:dataset}
The dataset we used is based on LASA dataset~\cite{khansari2011learning} which consists of $30$ shapes, each with $7$ trajectories of $1000$ points in $2$- dimensional. We recombined the dimensions of different trajectories of the same shape and got $4$ $3$-dimensional trajectories for each shape.
More in detail, we first stacked the $7$ demonstrations of each shape in a matrix $\bm{D}_i$ for $i=1,\ldots,30$ with $14$ rows and $1000$ columns. Given $\bm{D}_i$, we extracted the $4$ demonstrations by selecting the rows $[0,1,2]$, $[4,5,6]$, $[8,9,10]$, and $[12,3,0]$. In this way, the obtained $4$ demonstrations have the third dimension sampled from the $x$-axis of the original data. This implies that the added dimension contains similar patterns for the same shape, as usually required in \ac{lfd}.

\begin{figure}[!t]
	\centering
	\def\svgwidth{.8\linewidth}
	{\fontsize{8}{8}
\begingroup%
  \makeatletter%
  \providecommand\color[2][]{%
    \errmessage{(Inkscape) Color is used for the text in Inkscape, but the package 'color.sty' is not loaded}%
    \renewcommand\color[2][]{}%
  }%
  \providecommand\transparent[1]{%
    \errmessage{(Inkscape) Transparency is used (non-zero) for the text in Inkscape, but the package 'transparent.sty' is not loaded}%
    \renewcommand\transparent[1]{}%
  }%
  \providecommand\rotatebox[2]{#2}%
  \newcommand*\fsize{\dimexpr\f@size pt\relax}%
  \newcommand*\lineheight[1]{\fontsize{\fsize}{#1\fsize}\selectfont}%
  \ifx\svgwidth\undefined%
    \setlength{\unitlength}{200.25bp}%
    \ifx\svgscale\undefined%
      \relax%
    \else%
      \setlength{\unitlength}{\unitlength * \real{\svgscale}}%
    \fi%
  \else%
    \setlength{\unitlength}{\svgwidth}%
  \fi%
  \global\let\svgwidth\undefined%
  \global\let\svgscale\undefined%
  \makeatother%
  \begin{picture}(1,0.51310861)%
    \lineheight{1}%
    \setlength\tabcolsep{0pt}%
    \put(0.02992376,0.23316387){\color[rgb]{0,0,0}\makebox(0,0)[lt]{\lineheight{1.25}\smash{\begin{tabular}[t]{l}$y$\end{tabular}}}}%
    \put(0.26968284,0.0081983){\color[rgb]{0,0,0}\makebox(0,0)[lt]{\lineheight{1.25}\smash{\begin{tabular}[t]{l}$x$\end{tabular}}}}%
    \put(0.70283512,0.0081983){\color[rgb]{0,0,0}\makebox(0,0)[lt]{\lineheight{1.25}\smash{\begin{tabular}[t]{l}$x$\end{tabular}}}}%
    \put(0,0){\includegraphics[width=\unitlength,page=1]{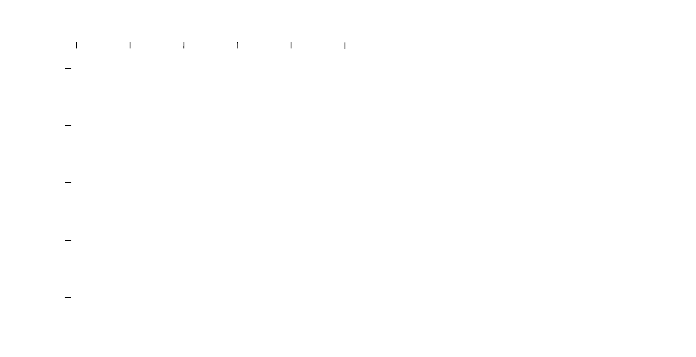}}%
    \put(0.93386225,0.26347449){\color[rgb]{0,0,0}\makebox(0,0)[lt]{\lineheight{1.25}\smash{\begin{tabular}[t]{l}$y$\end{tabular}}}}%
    \put(0.92866367,0.04075414){\color[rgb]{0,0,0}\makebox(0,0)[lt]{\lineheight{1.25}\smash{\begin{tabular}[t]{l}$z$\end{tabular}}}}%
    \put(0,0){\includegraphics[width=\unitlength,page=2]{ori_new_dataA2.pdf}}%
  \end{picture}%
\endgroup%
}
	\caption{\emph{Left}: ``A'' shape of original LASA data. \emph{Right}: ``A'' shape of created data.}
	\label{fig:ori_new_dataA}
\end{figure}

One example of ``A'' shape data is shown in Fig.~\ref{fig:ori_new_dataA}. We consider those 3D data as projections in the tangent space of \ac{uq} from which one can directly compute \ac{uq} profiles using~\eqref{uq_exp}. For \ac{spd} matrices, we assume that the 3D data represent the vectorization of symmetric $2\times 2$ matrices obtained with Mandel's notation. We then invert  Mandel's notation to compute $2 \times 2$ symmetric matrices (in tangent space) and retrieve \ac{spd} matrices using the exponential mapping in~\eqref{spd_exp}. The center of the tangent space was placed at $\bm{g}_\text{SPD} = \mathrm{diag}([100, 100])$ for SPD matrices and at $\bm{g}_\text{q}$ = $1 + [0, 0, 0]$ for \acp{uq}.
Before training, we normalized each dimension of the tangent space data to zero mean and unit variance. 

\subsection{Training process}
Being each shape significantly different (see Fig.~\ref{fig:Total}), we trained different models for each of the $30$ shapes in the dataset. We randomly chose several shapes of our dataset to test the performance of the trained model. For each shape, the corresponding  $4$ demonstrations were used as the dataset for training.

\begin{figure}[!t]
	\centering
	\def\svgwidth{1\linewidth}
	{\fontsize{6}{6}
		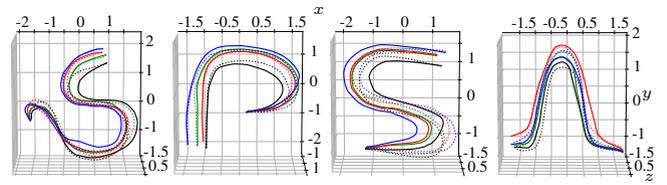}
	\caption{Visualization of data generated by our model (solid line) and given demonstrations (dot line) on the tangent space.}
	\label{fig:4trajectories}
\end{figure}

In this stage, the number of coupling layers with random initialization was $10$, playing the role of the emission function. Generally, after $40$ epochs of training, the generated trajectories could reproduce accurately the given pattern. Further, all trajectories converged to the goal point in the end, which empirically proves the stability of the model. Examples of these data are shown in Fig.~\ref{fig:4trajectories}, where all the trajectories were generated by the model trained for only $40$ epochs.

During the training process, \ac{dtw} \cite{muller2007dynamic}, indicating the similarity between the generated trajectories and given demonstrations, was considered as the evaluation factor and it was computed using the implementation provided by~\cite{Jekel2019}. We monitored the value changes of \ac{dtw} during $200$ epochs of training for all $30$ shapes, the value reached the lowest before $100$ epochs in $25$ cases out of $30$. Therefore, for hyperparameter search, the maximum number of training epoch was limited to $100$ to accelerate the process.
\hlnew{
Typical loss curves trained for $200$ epochs are shown in \mbox{Fig.~\ref{fig:TrainingCurves}} .
}

\subsection{Hyperparameters search}\label{subsec:hyper search}
There are several hyperparameters in our model, including the number of layers, the activation function, the learning rate, and the optimizer. But before searching for the best combination, we first tested some different initialization methods to make sure the inside parameters were set correctly in the beginning. We tried popular initialization methods like
He~\etal~\cite{he2015delving} and found out that randomly uniform initialization made the training process to converge faster. Therefore, we used it in the following search process.

To make sure that different hyperparameter combinations affecting on the same model we set a certain seed for random initialization and chose the data of ``N'' shape to train the model. The search method used is provided by Optuna package~\cite{optuna_2019} and it automatically prunes improper combinations based on the evaluation values during the search and generates the combination that tends to achieve a better performance, which can accelerate the whole process significantly. 

During the first rough search, the options for the hyperparameters are shown in Table~\ref{tab:initial_hyper}. We generated $100$ combinations among which only $37$ combinations were executed completely and the rest were pruned. From the search results, we noticed that for the activation function ReLu beats the other one; for the learning rate, the range shrinks to 1e-4 to 1e-2; and for the optimizer, Adam and Adamax~\cite{kingma2014adam} stand out. Also, we found Adamax was able to generate more stable training processes while Adam tended to reach lower minimum of errors.

\begin{table}[t]
	\begin{center}
		\caption{Hyperparameters considered in the search, their initial range, and the best value found.}
		\label{tab:initial_hyper}
		\begin{tabular}{|l|l|l|}
			\hline
			\sc{Parameter}& \sc{Options} & \sc{Best value}                      \\ 
			\hline
			number of   layers    & 8 to 12 & 11                      \\
			activation   function & ReLu, Tanh & ReLu                  \\
			optimizer             & Adam, Adamax,   SGD, RMSprop & Adam\\
			learning rate         & 1e-5 to 1e-1 & 0.00098\\
			\hline         
		\end{tabular}
	\end{center}
\end{table}

\begin{figure}[!t]
	\centering
	\def\svgwidth{1\linewidth}
	{\fontsize{8}{8}
		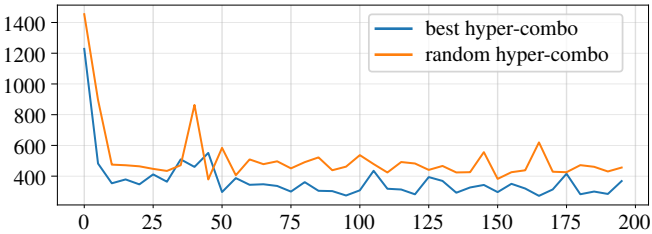}
	\caption{\hlnew{Loss curves indicating training process for ``N'' shape dataset. Loss is computed every 5 epochs. The orange curve indicates the training process under a random hyperparameter combination while the blue curve shows the loss trend with the best combination searched in \mbox{Sec.\ref{subsec:hyper search}}. }  }
	\label{fig:TrainingCurves}
\end{figure}

Based on the above results, we carried out several rounds of fine search only utilizing ReLu as the activation function and just $2$ options (Adam or Adamax) for the optimizer and the shrunk range for learning rate. However, since the random initialization shall also affect the training process, we gave different random seeds for different search processes. With this procedure and a seed of $20$, we found the best hyperparameter configuration in Tab.~\ref{tab:initial_hyper} resulting in a \ac{dtw} error of $274$. \hlnew{\mbox{Figure.~\ref{fig:TrainingCurves}} shows the learning curve with and without hyperparameters search. Moreover, the figure shows that the lower training error is achieved with the best parameters.
}

Regarding training and generation time, using the best hyperparameter setting in Tab.~\ref{tab:initial_hyper}, a batch size of $128$ and $100$ training epochs, and considering that a motion trajectory contains $4000$ samples ($3$D vector), the average training time is $61$ minutes for one shape utilizing a NVIDIA RTX 2060 GPU. Our approach takes only $20\,$s to generate an entire trajectory of $1000$ samples.

For the best hyperparameters, we did further testing by transferring the generated data on the tangent space back to the manifold ($\mathcalbold{S}^{+}_2$ and $\mathcalbold{S}^3$) and comparing the distance between generated and given data. For \ac{spd} matrices, we use the \ac{led}\cite{PENNEC202075}:
\begin{equation}{
\mathrm{LEd} = \left \| \mathrm{logm}(\bm{s})-\mathrm{logm}(\hat{\bm{s}}) \right \| }_{F} , \label{eq}\end{equation}
where $\bm{s}$ and $\hat{\bm{s}}$ are the given and generated \ac{spd} matrices  respectively, $\mathrm{logm}(\cdot)$ is the matrix logarithm, and $\Vert \cdot \Vert_F$ indicates the Frobenius Norm.

For \acp{uq}, we use the \ac{lqd}:
\begin{equation}{
\mathrm{LQd}=\left \| 2\Logg(\langle \bm{q}, \overline{\hat{\bm{q}}} \rangle) \right \| } , 
	\label{eq}\end{equation}
where $\bm{q}$ is the given quaternion, $\overline{\hat{\bm{q}}}$ is the conjugate of the generated quaternion, and $\langle \cdot,\cdot \rangle$ represents the Hamilton product, whose result is a quaternion mapped to the tangent space by the ~\eqref{uq_log} to be a vector.

\begin{figure}[!t]
	\centering
    \includegraphics[width=\columnwidth]{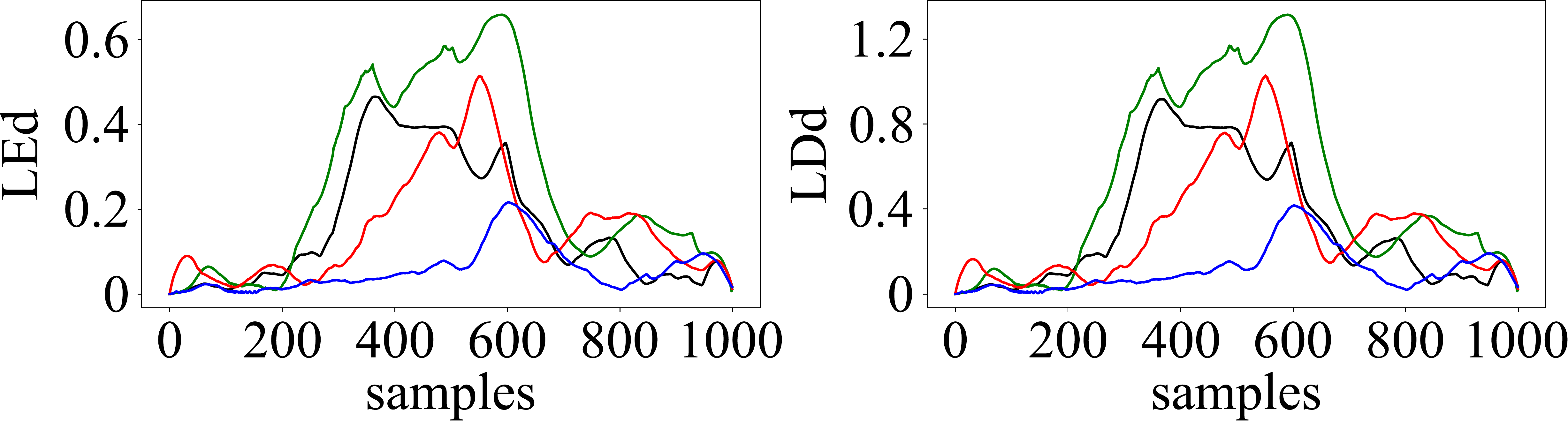}
	\caption{Distances of each pair of SPD matrix and quaternion.}
	\label{fig:2error}
\end{figure}

For each demonstration, we also visualized the change of the \ac{led} and \ac{lqd} along the sequence in Fig.~\ref{fig:2error}. From which we notice that the distance error in the middle of the set are larger than that of the start and the end. The reason is that the $4$ starting samples are used to generated the whole $4$ sets, and because of the stability of the model, in the end, they will converge to the last sample. Hence, the distances in the start and the end are very small. In the middle of the motion, the generated data depend on the pattern that the model has learned, and the error also accumulates there, so the distance reaches its highest in the middle. It is worth mentioning that the shape of the $2$ graphs in in Fig.~\ref{fig:2error} is the same because they all come from the same data on the tangent space.


\subsection{Generality testing}\label{subsec:generality}
The top $3$ hyperparameter combinations, found using the approach described in the previous section, were computed considering only data of the shape ``N''. In this section, we investigate whether they are also proper for training models for other shapes. Therefore, we took the combination with the best performance to train the other shapes and compared the results with those from random combination but also with $11$ layers (making sure that the $2$ models have the same capacity).
%

\begin{figure}[!t]
	\centering
	\def\svgwidth{0.9\linewidth}
	{\fontsize{8}{8}
\begingroup%
  \makeatletter%
  \providecommand\color[2][]{%
    \errmessage{(Inkscape) Color is used for the text in Inkscape, but the package 'color.sty' is not loaded}%
    \renewcommand\color[2][]{}%
  }%
  \providecommand\transparent[1]{%
    \errmessage{(Inkscape) Transparency is used (non-zero) for the text in Inkscape, but the package 'transparent.sty' is not loaded}%
    \renewcommand\transparent[1]{}%
  }%
  \providecommand\rotatebox[2]{#2}%
  \newcommand*\fsize{\dimexpr\f@size pt\relax}%
  \newcommand*\lineheight[1]{\fontsize{\fsize}{#1\fsize}\selectfont}%
  \ifx\svgwidth\undefined%
    \setlength{\unitlength}{636.12929637bp}%
    \ifx\svgscale\undefined%
      \relax%
    \else%
      \setlength{\unitlength}{\unitlength * \real{\svgscale}}%
    \fi%
  \else%
    \setlength{\unitlength}{\svgwidth}%
  \fi%
  \global\let\svgwidth\undefined%
  \global\let\svgscale\undefined%
  \makeatother%
  \begin{picture}(1,0.28092672)%
    \lineheight{1}%
    \setlength\tabcolsep{0pt}%
    \put(0,0){\includegraphics[width=\unitlength,page=1]{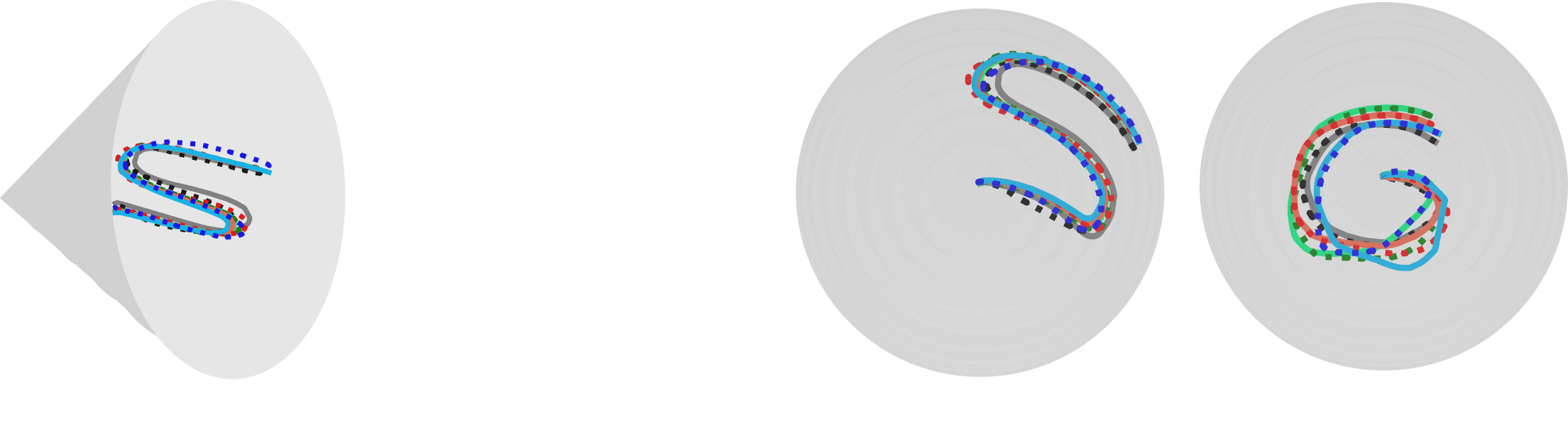}}%
    \put(0.11611507,0.00569239){\color[rgb]{0,0,0}\makebox(0,0)[lt]{\lineheight{1.25}\smash{\begin{tabular}[t]{l}(a)\end{tabular}}}}%
    \put(0.37492065,0.00569239){\color[rgb]{0,0,0}\makebox(0,0)[lt]{\lineheight{1.25}\smash{\begin{tabular}[t]{l}(b)\end{tabular}}}}%
    \put(0.60156149,0.00569239){\color[rgb]{0,0,0}\makebox(0,0)[lt]{\lineheight{1.25}\smash{\begin{tabular}[t]{l}(c)\end{tabular}}}}%
    \put(0.86087877,0.00569239){\color[rgb]{0,0,0}\makebox(0,0)[lt]{\lineheight{1.25}\smash{\begin{tabular}[t]{l}(d)\end{tabular}}}}%
    \put(0,0){\includegraphics[width=\unitlength,page=2]{cone_sphere2.pdf}}%
  \end{picture}%
\endgroup%
}
	\caption{Two different shapes visualized on the respective \acp{rm}, \ie $\mathcalbold{S}^{+}_2$ ((a) and (b)) and $\mathcalbold{S}^3$ ((c) and (d)). Light solid (dark dotted) lines represent the generated (demonstrated) data.}
	\label{fig:cone_sphere}
\end{figure}

We experimentally found out that the average \ac{dtw} error ($\mathrm{DTW_e}$) of the whole dataset given by the best hyperparameter combination of ``N'' shape ($\mathrm{mean~DTW_e} = 284$) is slightly better than that by random combination ($\mathrm{mean~DTW_e} = 294$). Although it is not what we expected but it is also reasonable since the patterns to be learned by the model are distinguishing, the best combination for one pattern maybe just a random one for another pattern. As qualitatively shown in Fig.~\ref{fig:cone_sphere}, where we visualized two sample motions on their manifolds, the learning process and the successive projection of the generated profiles on the corresponding \ac{rm} is already accurate. Indeed the projection's analytical form is know and does not affect the accuracy. If the particular task demands for higher reproduction accuracy, it is recommended to search for a better combination of hyperparameters in the specific case. For instance, we observed in the case of ``N'' shape that the \ac{dtw} error decreases from $378.5$ to $274$, \ie by $27.6\,$\% with the best set of hyperparameters.

\begin{figure}[!t]
	\centering
	\includegraphics[width=\linewidth]{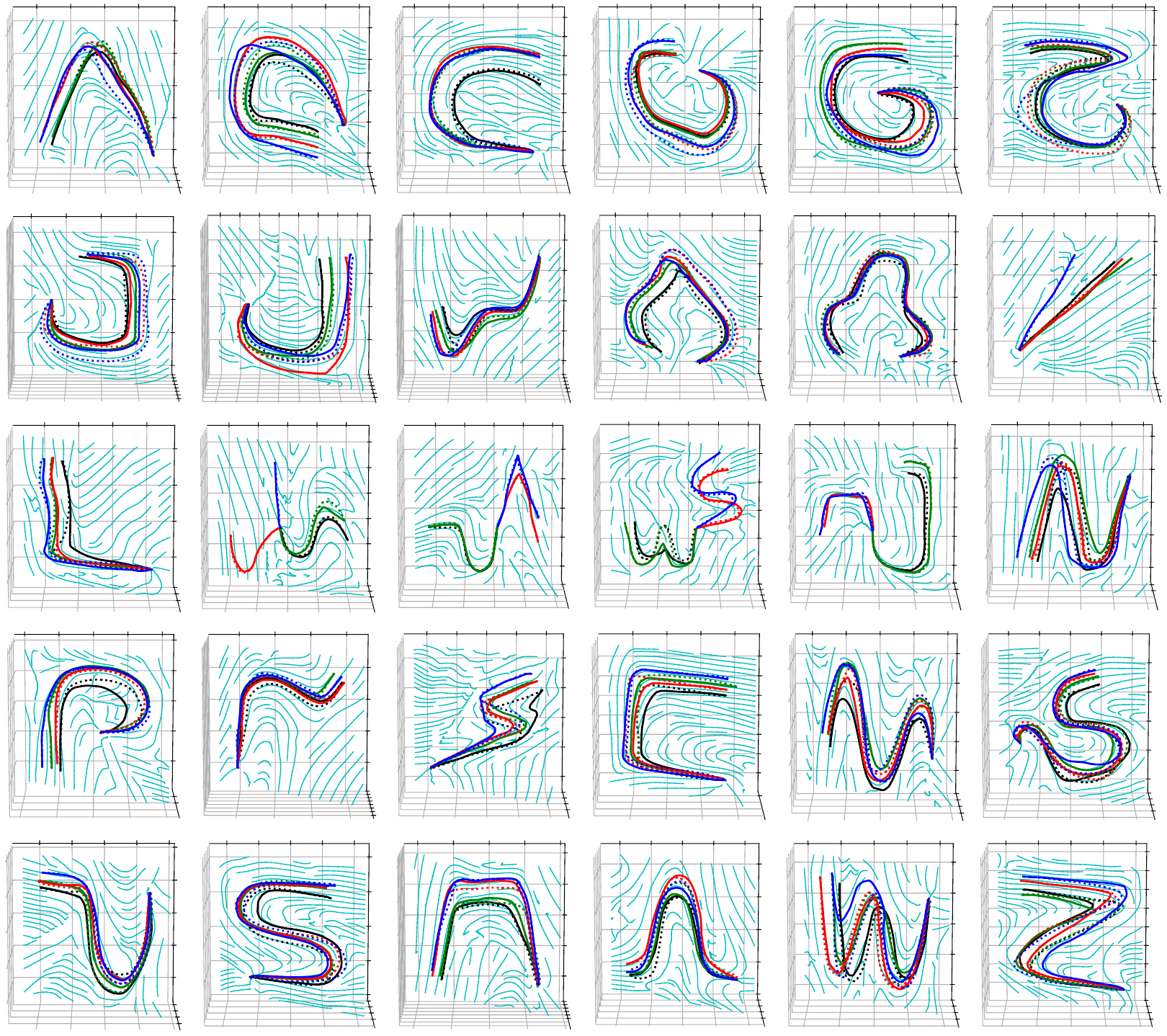} 
	\caption{Stream figures of the demonstration area for each shape.}
	\label{fig:Total}
\end{figure}

Next, we also want to know the generality of the model concerning the demonstration area \ie another perspective of the robustness of the learning results. To have a intuitive view of the model, we plotted the stream figures in the cubes containing the normalized data of demonstrations on the tangent space shown in Fig.~\ref{fig:Total}. We plotted the generated trajectories, the corresponding demonstrations, and the stream lines (cyan curves) representing the velocity field (direction to the next predicted point). To provide a better visualization without all stream curves in the area messing together, we just selected the plane most fitting those demonstrations to show the shape patterns that the model has learned.

From these stream figures, the learning results of the model are well illustrated. First, it is clear that the models are able to accurately learn the patterns of very different shapes. Second, except the area that the demonstrations go through, the model also equips the surrounding space with the similar basic pattern which also proves the robustness of the model. Last, all the generated stream curves tend to converge to the final point of the demonstrations indicating the stability of the model.

\subsection{Comparison}\label{subsec:simulation}
We compare our approach with Riemannian \ac{gmm} (R-\ac{gmm})~\cite{Calinon20RAM} and \ac{fdm}~\cite{perrin2016fast}, as well as with two baseline approaches. R-\ac{gmm} extends the classic \ac{gmm}/\ac{gmr} appraoch to perform regression on \ac{uq} and \ac{spd} manifolds. \ac{fdm} has been proposed to learn a diffeomorphism between Euclidean spaces. The procedure presented in Sec.~\ref{sec:method} allows to extend \ac{fdm} to learn stable Riemannian skills. 
The baselines are naive versions of our procedure where we ignore the constraints from the manifold during the learning and post-process the generated data in order to fulfill the geometric constraints. In particular, unit quaternions are considered as $4$-D Euclidean vectors during the learning and normalized after the generation. \ac{spd} matrices are vectorized by stacking the $3$ unique components of the $2\times 2$ \ac{spd} matrix into a $3$-D Euclidean vector used for learning\footnote{As \ac{spd} matrices are symmetric matrices, then we use Mandel's notation to vectorize the \ac{spd} matrices, and use the inverse of Mandel's notation to matricize an Euclidean vector.}. From the generated vector, we build a symmetric matrix and find the nearest \ac{spd} matrix to this one using the approach in~\cite{Higham1988}. \hlnew{For all the approaches, we train a separate model for each manifold (\mbox{\ac{uq}} and \mbox{\ac{spd}}) and each motion in the Riemannian LASA dataset (Sec.~\mbox{\ref{subsec:dataset}}). For validation, we take only the first point in each demonstration and use it to generate the entire trajectory of $1000$ steps as described in Algorihm~\mbox{\ref{alg:riem_flow}} (point 3). This procedure is commonly adopted to validate LfD approaches based on stable dynamical systems~\mbox{\cite{khansari2011learning,perrin2016fast,UrainJulen2020ILDS}}.}

\begin{figure}[!t]
	\centering
	\includegraphics[width=\columnwidth]{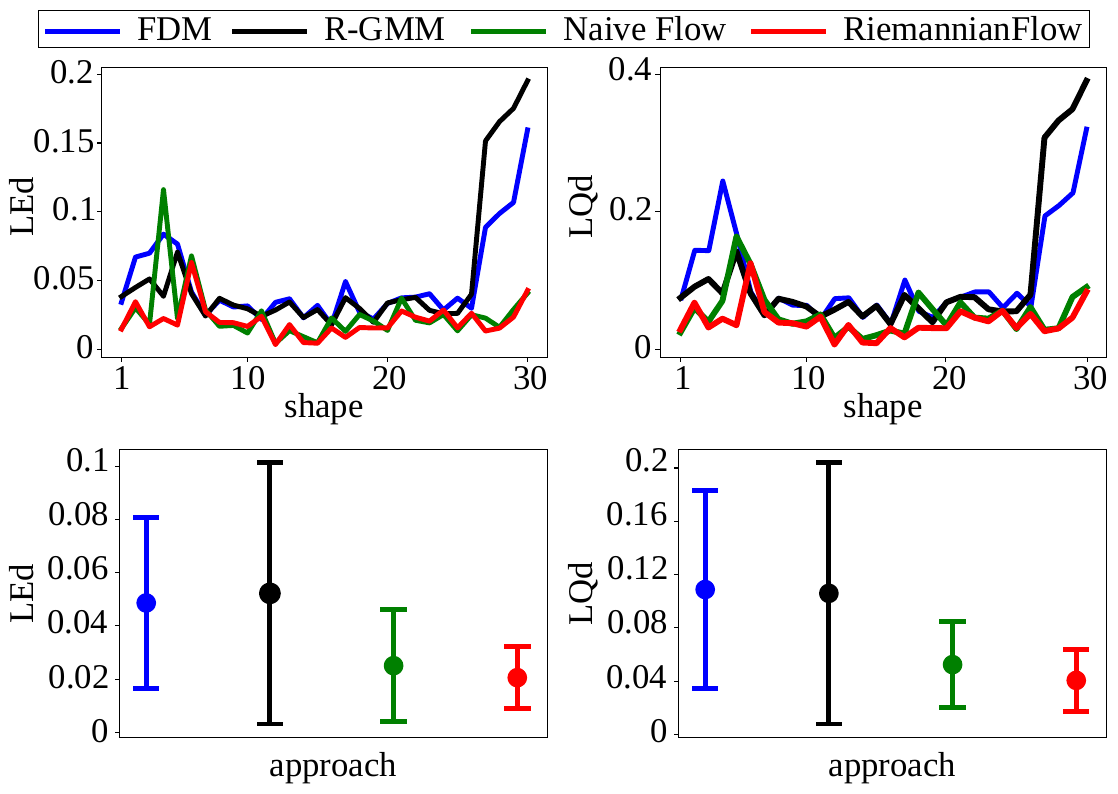} 
	\caption{Comparative results obtained with different approaches on the Riemannian LASA dataset for SPD matrix (left \hlnew{column}) and \ac{uq} (right \hlnew{column}).}
	\label{fig:comparison}
\end{figure}

From the comparison results shown in Fig.~\ref{fig:comparison},  transferring the data to the tangent space and acquiring the corresponding model will lead better \hlnew{average} performance (8.2\% for \ac{spd} and 17.7\% for \ac{uq} on average).  \hlnew{The bottom row in Fig.~\mbox{\ref{fig:comparison}} shows mean accuracy and standard deviation for each approach on the entire dataset. On average, RiemannianFlow outperforms all the considered approaches. Moreover, compared to the naive version, RiemannianFlow has similar performance, but more accurate as it has smaller mean and standard deviation. These results for \mbox{\ac{spd}} matrices are in-line with a previous study~\mbox{\cite{abu2018force}}, where authors have shown that the approximation error becomes significant for higher dimensional matrices.}
Moreover, looking at the shapes $27$ to $30$ in Fig.~\ref{fig:comparison} \hlnew{(first row)}, R-\ac{gmm} and \ac{fdm} fail in learning \textit{multimodal} skills with different behaviors in different parts of the manifold (see the third row, columns $2$ to $5$ of Fig.~\ref{fig:Total}). This is because \ac{fdm} learns from a single (average) demonstration, while R-\ac{gmm} takes a time-like input and regress always the same motion on the manifold. On the contrary, RiemannianFlow is capable to accurately encode single- and multimodal Riemannian skills.

\subsection{Robot experiment}\label{subsec:real_experiment}
In order to evaluate our RiemannianFlow approach experimentally, we studied a typical industrial insertion task, namely  \acl{pih}.
However, we applied it to a shape fitting toy. In this game, the robot needs to follow accurately the demonstrated trajectory to perform a successful insertion. It is worth mentioning that, even if several state-of-the-art methods exist to solve PiH, we used PiH as an industrial application example as it shows the key features of our procedure.

\begin{figure}[t]
	\centering
	\def\svgwidth{1\linewidth}
	{\fontsize{8}{8}
\begingroup%
  \makeatletter%
  \providecommand\color[2][]{%
    \errmessage{(Inkscape) Color is used for the text in Inkscape, but the package 'color.sty' is not loaded}%
    \renewcommand\color[2][]{}%
  }%
  \providecommand\transparent[1]{%
    \errmessage{(Inkscape) Transparency is used (non-zero) for the text in Inkscape, but the package 'transparent.sty' is not loaded}%
    \renewcommand\transparent[1]{}%
  }%
  \providecommand\rotatebox[2]{#2}%
  \newcommand*\fsize{\dimexpr\f@size pt\relax}%
  \newcommand*\lineheight[1]{\fontsize{\fsize}{#1\fsize}\selectfont}%
  \ifx\svgwidth\undefined%
    \setlength{\unitlength}{611.66565982bp}%
    \ifx\svgscale\undefined%
      \relax%
    \else%
      \setlength{\unitlength}{\unitlength * \real{\svgscale}}%
    \fi%
  \else%
    \setlength{\unitlength}{\svgwidth}%
  \fi%
  \global\let\svgwidth\undefined%
  \global\let\svgscale\undefined%
  \makeatother%
  \begin{picture}(1,0.41784203)%
    \lineheight{1}%
    \setlength\tabcolsep{0pt}%
    \put(0,0){\includegraphics[width=\unitlength,page=1]{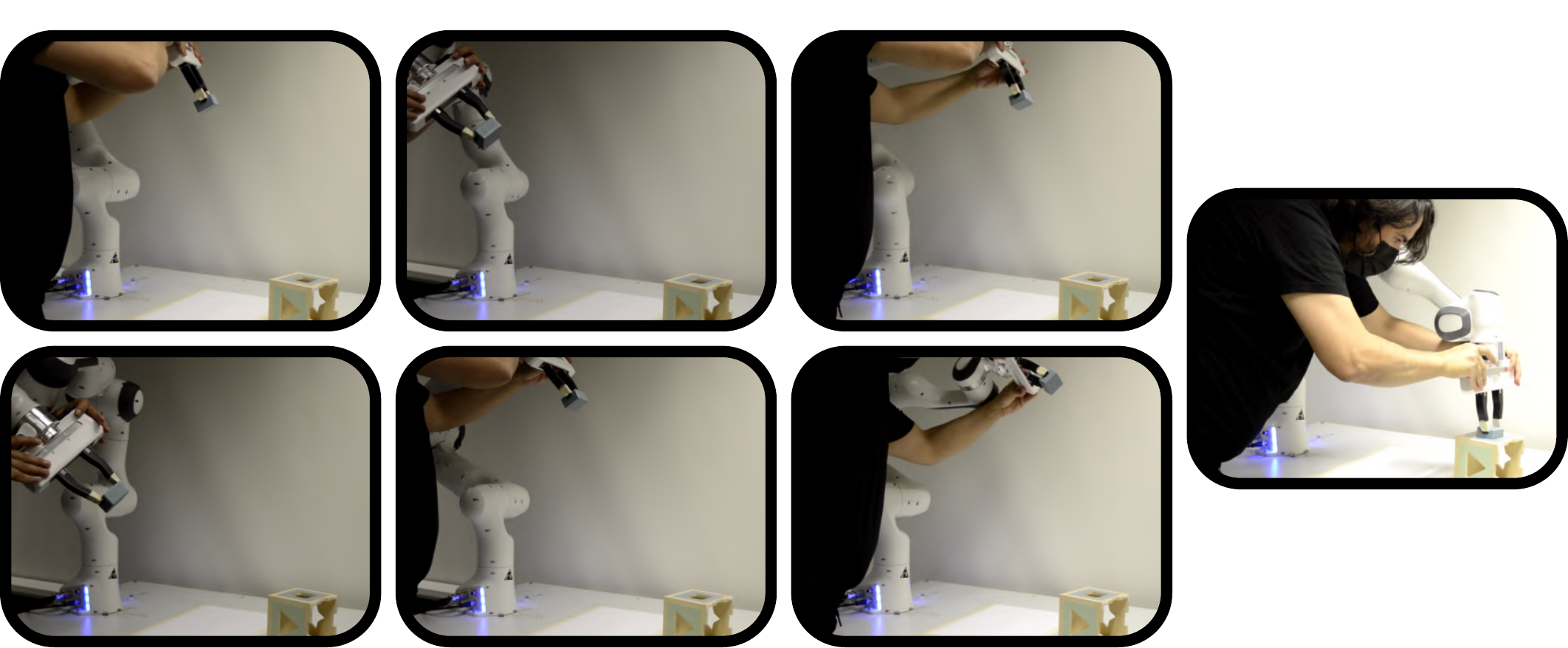}}%
    \put(0.81653246,0.31046902){\color[rgb]{0,0,0}\makebox(0,0)[lt]{\lineheight{1.25}\smash{\begin{tabular}[t]{l}Goal pose\end{tabular}}}}%
    \put(0.07991407,0.40649046){\color[rgb]{0,0,0}\makebox(0,0)[lt]{\lineheight{1.25}\smash{\begin{tabular}[t]{l}Demonstrations from different starting poses\end{tabular}}}}%
  \end{picture}%
\endgroup%
}
	\caption{An illustration of the \ac{pih} experiment. A human operator is teaching the robot to perform \ac{pih} task by starting from different poses and ending up in the hole pose.}
	\label{fig:realex}
\end{figure}

\begin{figure}[!t]
	\centering
	\def\svgwidth{1\linewidth}
	{\fontsize{8}{8}
		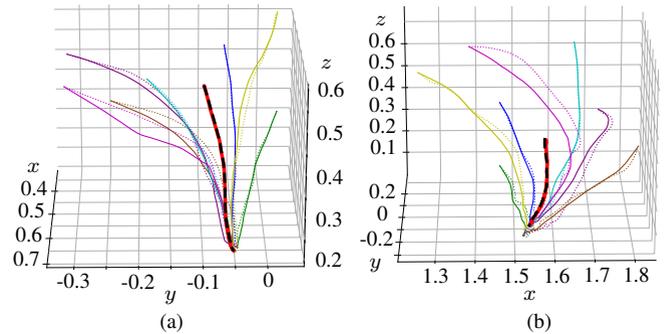}
	\caption{Results for the \ac{pih} task. (a) Position and (b) orientation (projected on the tangent space) data of the real robot experiment. The dash black curves represent the prediction for a new starting point, and the thick red lines indicate the movement that robot performs in both figures.}
	\label{fig:real}
\end{figure}

We provided $7$ kinesthetic demonstrations starting at different poses and converging to the hole pose (see Fig.~\ref{fig:realex}) defined by the position of $\bm{g}_{\text{p}} = [0.675, -0.051, 0.203]\trsp\,$m and the orientation of \ac{uq}: $\bm{g}_{\text{q}} = 0.022 + [0.991, -0.125, 0.035]\trsp$. The demonstrations are in the form of $\{\{ \bm{p}^{demo}_{m,n}, \bm{q}^{demo}_{m,n} \}_{m=1}^{M}\}_{n=1}^{N}$ where $M=4000$ is the total length of the demonstrations and $N=7$ is the number of demonstrations.
\ac{ros} was used to record the demonstration data. Afterwards, we projected the \ac{uq} trajectories, from all demonstrations, to the tangent space $\mathcalbold{T}_{\bm{g}}\mathcalbold{S}^3$ using~\eqref{uq_log}.
Subsequently, we trained two models for both position and orientation trajectories. These 2 models were used later to predict a new pose trajectory that started from a new arbitrary pose, which was different from the demonstrations starting poses, and ended up in the hole pose performing a successful \ac{pih} insertion using \ac{ros} with the generated prediction data. An illustration of the demonstrations, the prediction (black dashed line), and robot pose  (red solid line) trajectories when performing in the tangent space is shown in Fig.~\ref{fig:real}.

In this experiment we did not performed an hyperparameter search and used the combination of ($11$ layers, ReLu, Adam, $0.001$) with random uniform initialization.
From Fig.~\ref{fig:real}, although there were situations where the generated trajectories were not perfectly following the given demonstrations, the convergence of the final state was ensured and the task was successfully executed. Moreover, when the robot performed the task according to a new generated trajectory, it also reached the same goal as expected. More executions of the task under different conditions is shown in the accompanying video.


\section{\hlnew{Discussion}}\label{sec:discussion}
\hlnew{
	Experiments in Sec.~\mbox{\ref{subsec:simulation}} and \mbox{\ref{subsec:real_experiment}} show the effectiveness of RiemannianFlow both in simulations and in a real experiment. The experimental comparison, performed on a public benchmark, has confirmed known results and highlighted some new findings. An expected result is that approaches that effectively learn from multiple demonstrations are in general more accurate. This can be clearly seen in the bottom plots of Fig.~\mbox{\ref{fig:comparison}} where both Naive and RiemannianFlow outperform FDM and R-GMM. Another expected results is data normalization affects the accuracy, especially if the motion is integrated over time as the error accumulate. In our comparison, each motion lasts for 1000 steps and the error is still limited (although already visible in Fig.~\mbox{\ref{fig:comparison}}). It is clear that, in a real setting where the motion may last for several minutes, the accumulated error may cause the task to fail. Similar considerations apply to the Cholesky decomposition, where increasing the matrix size also affects accuracy~\mbox{\cite{abu2018force}}. Finally, from our comparison, it is possible to conclude that imitation learning approaches based on dynamical systems are sufficiently flexible to accurately encode complex manifold motions while preserving the stability.  

    The accuracy of RiemannianFlow comes at the cost of a longer training time (Sec.~\mbox{\ref{subsec:hyper search}}), but in most cases, training can be performed in about 1 hour on a PC equipped with a consumer GPU---in our test where we used a NVIDIA RTX 2060 and acquire a good model. As for the hyperparameter search, an elaborate search will bring large gain of performance but cost longer time. And the search needs to be repeated to learn motions that significantly differ from each other (Sec.~\mbox{\ref{subsec:generality}}). Once the network is trained, the time needed to generate one sample allows to close a control loop at about \mbox{$50\,$}Hz (Sec.~\mbox{\ref{subsec:hyper search}}), which is sufficient for a kinematic controller. Therefore, RiemannianFlow can be potentially deployed on resource-constrained robots like mobile platforms, exploiting low-power GPUs like the NVIDIA Jetson series.
}

\section{Conclusion and Future work}\label{sec:conclusion}
In this paper, have we proposed a deep generative model to properly encode complex data that have to fulfil specific geometric constraints. From the geometric perspective, we consider those constrained data forming \acp{rm}, and utilize distance reserving mappings to project them on the tangent space, successfully releasing the constrains. After the learning in the tangent space, we project back the generated data on the manifold to re-fulfill the original constraints.

The resulting approach, namely RiemannianFlow, has been evaluated on a benchmark of Riemannian motions and in a \ac{pih} task with a real robot. Obtained results show that the approach has the ability of learning various kinds of complex Riemannian patterns while guaranteeing the stability and fulfilling geometric constraints for both stiffness (\ac{spd} matrices) and orientation data (\acp{uq}). Although the learned model is already accurate if the learning process starts with a random initialization of the hyperparameters, hyperparameters search has been shown to be an effective solution to significantly reduce the reproduction error, which is necessary for for tasks that require higher accuracy (\eg \ac{pih} task). Last, through stream figures, we prove the high robustness of the model in a region around the demonstration area.

In the future work, we will try to let the model learn more control information including position, orientation, stiffness, and velocity at the same time, increasing the dependence among those attributes to handle more sophisticated situations. Further, we will find some faster initialization methods to give better performance in general.


\bibliographystyle{IEEEtran}
\bibliography{ref}
\begin{IEEEbiography}[{\includegraphics[width=1in,height=1.25in,clip,keepaspectratio]{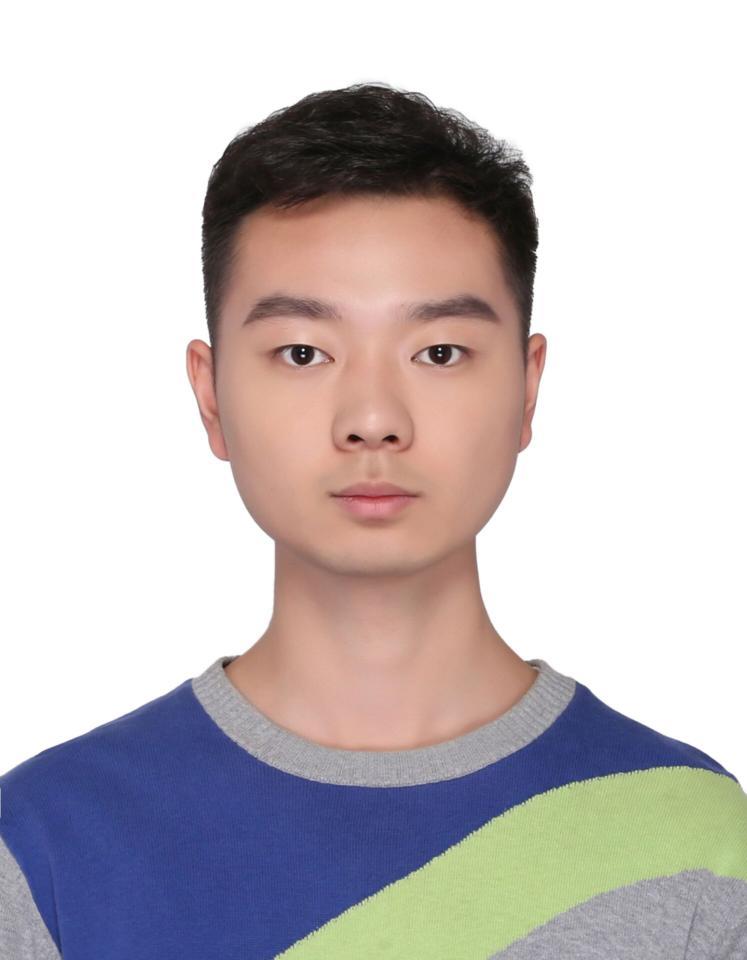}}]{Weitao Wang} received the B.S. degree in automation engineering from University of Electronic Science and Technology of China (UESTC), Chengdu, China in 2020, and a master in 2022 in Autonomous System Programme of European Institute of Innovation \& Technology (EIT), a joint program between KTH Royal Institute of Technology, Stockholm, Sweden and Aalto University, Espoo, Finland. Besides, he works as a research assistant at Intelligent Robotics Group at EEA, Aalto University since June, 2021.
\end{IEEEbiography}

\begin{IEEEbiography}[{\includegraphics[width=1in,height=1.25in,clip,keepaspectratio]{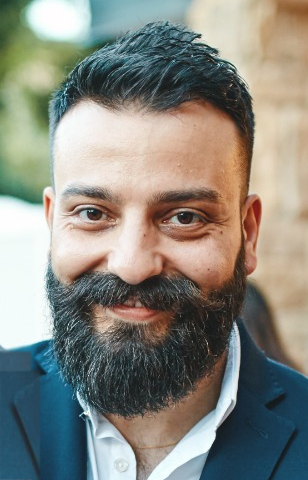}}]{Matteo Saveriano} received his B.Sc. and M.Sc. degree in automatic control engineering from University of Naples, Italy,  in 2008 and 2011, respectively. He received is Ph.D. from the Technical University of Munich in 2017. 	Currently, he is an assistant professor at the Department of Industrial Engineering (DII), University of Trento, Italy. Previously, he was  an assistant professor at the University of Innsbruck and a post-doctoral researcher at the German Aerospace Center (DLR). He is an Associate Editor for RA-L. His research activities include robot learning, human-robot interaction, understanding and interpreting human activities. Webpage: https://matteosaveriano.weebly.com/ 
\end{IEEEbiography}

\begin{IEEEbiography}[{\includegraphics[width=1in,height=1.25in,clip,keepaspectratio]{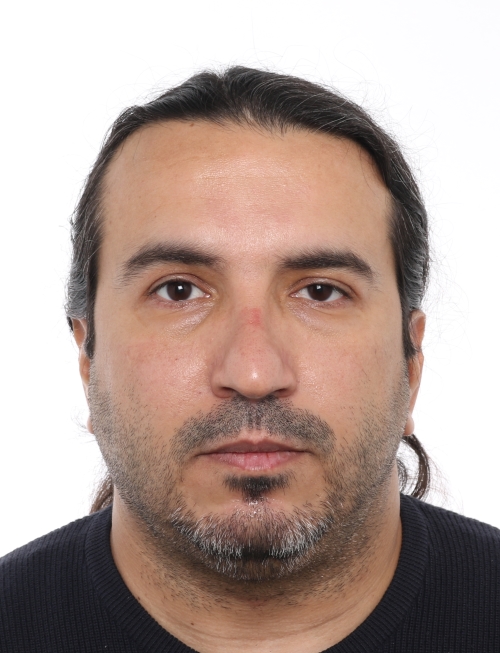}}]{Fares J. Abu-Dakka} received his B.Sc. degree in mechanical engineering from Birzeit University, Palestine, in 2003, and his M.Sc. and Ph.D. degrees in robotics motion planning from the Polytechnic University of Valencia, Spain, in 2006 and 2011, respectively. 	Currently, he is a senior researcher at Intelligent Robotics Group at EEA, Aalto University, Finland. Previously, he was researching at ADVR, Istituto Italiano di Tecnologia (IIT). Between 2013 and 2016 he was holding a visiting professor position at ISA of the Carlos III University of Madrid, Spain. He is an Associate Editor for ICRA, IROS, and RA-L. His research activities include robot learning \& control, human-robot interaction. Webpage: https://sites.google.com/view/abudakka/ 
\end{IEEEbiography}

\EOD
\end{document}